\documentclass{ieeeaccess}
\usepackage{cite}
\usepackage{amsmath,amssymb,amsfonts}
\usepackage{algorithmic}
\usepackage{graphicx}
\usepackage{textcomp}
\usepackage{algorithm}
\usepackage{adjustbox}
\usepackage{multirow}
\usepackage{xspace}
\usepackage{url}
\def\BibTeX{{\rm B\kern-.05em{\sc i\kern-.025em b}\kern-.08em
    T\kern-.1667em\lower.7ex\hbox{E}\kern-.125emX}}

\begin{document}
\history{Date of submission xxxx 00, 0000, date of current version xxxx 00, 0000.}
\doi{10.1109/ACCESS.2023.0322000}

\title{Enhancing Zero-Shot Crypto Sentiment with Fine-tuned Language Model and Prompt Engineering}
\author{\uppercase{RAHMAN S M WAHIDUR}\authorrefmark{1},
\uppercase{Ishmam Tashdeed}\authorrefmark{2}, 
\uppercase{Manjit Kaur}\authorrefmark{3}, \IEEEmembership{Senior Member, IEEE},
and HEUNG-NO-LEE
\authorrefmark{1},
\IEEEmembership{Senior Member, IEEE}}

\address[1]{School of Electrical Engineering and Computer Science, 
Gwangju Institute of Science and Technology, Gwangju 61005, South Korea}
\address[2]{Department of Computer Science and Engineering, 
Islamic University of Technology, Dhaka, Bangladesh.}
\address[3]{School of Computer Science and Artificial Intelligence, 
SR University, Warangal, Telangana, India-506371.}
\tfootnote{This work was supported by Institute of Information \& communications Technology Planning \& Evaluation (IITP) grant 
funded by the Korea government (MSIT) (No.2023-2021-0-00118, Development of decentralized consensus composition technology for 
large-scale nodes) and This research was supported by the MSIT(Ministry of Science and ICT), Korea, under the ITRC (Information Technology
Research Center) support program (IITP-2023-2021-0-01835) supervised by the IITP (Institute of Information \& Communications
Technology Planning \& Evaluation)}

\markboth
{Rahman. Wahidur \headeretal: Enhancing Zero-Shot Crypto Sentiment with Fine-tuned Language Model and Prompt Engineering}
{}

\corresp{Corresponding author: Heung-No Lee (heungno@gist.ac.kr).}

\begin{abstract}
Blockchain technology has revolutionized the financial landscape, with cryptocurrencies gaining widespread adoption for their decentralized and transparent nature. 
As the sentiment expressed on social media platforms can significantly influence cryptocurrency discussions and market movements, sentiment analysis has emerged as a crucial tool for understanding public opinion and predicting market trends. 
Motivated by the aim to enhance sentiment analysis accuracy in the cryptocurrency domain, this paper investigates fine-tuning techniques on large language models.
This paper also investigates the efficacy of supervised fine-tuning and instruction-based fine-tuning on large language models for unseen tasks. 
Experimental results demonstrate a significant average zero-shot performance gain of 40\% after fine-tuning, highlighting the potential of this technique in optimizing pre-trained language model efficiency. 
Additionally, the impact of instruction tuning on models of varying scales is examined, revealing that larger models benefit from instruction tuning, achieving the highest average accuracy score of 75.16\%. 
In contrast, smaller-scale models may experience reduced generalization due to the complete utilization of model capacity. 
To gain deeper insight about how instruction works with these language models, this paper presents an experimental investigation into the response of an instruction-based model under different instruction tuning setups. The investigation demonstrates that the model achieves an average accuracy score of 72.38\% for short and simple instructions. This performance significantly outperforms its accuracy under long and complex instructions by over 12\%, thereby effectively highlighting the profound significance of instruction characteristics in maximizing model performance.
Finally, this paper explores the relationship between fine-tuning corpus size and model performance, finding an optimal corpus size of 6,000 data points for achieving the highest performance across different language models. 
Significantly, a distillation model of BERT called MiniLM, published by the Microsoft research team, stands out for its exceptional data efficiency, effectively optimizing its performance while making efficient use of data. Conversely, Fine-tuned LAnguage Net, FLAN-T5 developed by the Google research team, impressively maintains consistent and reliable performance across diverse corpora, further affirming its robustness and versatility. 
\end{abstract}


\titlepgskip=-21pt

\maketitle

\section{Introduction}
\label{sec:introduction}
\PARstart{I}{n} the recent decade, cryptocurrency has gained much traction in finance and business for its decentralization, permissionless, and open nature built on the public blockchain \cite{Saad2018}.
This architecture can create an immutable and highly interoperable financial system with unprecedented transparency, equal access rights, and little need for central authorities that smart contracts can control \cite{Buterin2014}.
Cryptocurrencies employ cryptographic ciphers to facilitate financial transactions, distinguishing them from traditional forms of currency \cite{Shahbazi2021}.
Cryptocurrencies are the first pure digital assets to be included by asset managers \cite{Fang2022}.
They mitigate double-spending by leveraging multiple verifications from neighboring nodes within the blockchain network. 
As the number of confirmations grows, the transaction gains enhanced reliability and become highly irreversible. 
Because of these favorable attributes and the widespread accessibility of cryptocurrencies, they can be a means of transaction and a store of wealth \cite{Fang2022, Jay2020}.

In cryptocurrency discussions, social media networks have gained significant importance as information-sharing platforms. 
Communicating crypto events and market conditions through social media is widely recognized. 
Moreover, there is a prevalent belief in the correlation between altcoin prices and the sentiment expressed within the Twitter community \cite{Sasmaz2021}. 
The demand for cryptocurrency is intricately tied to people's trust in Bitcoin and its underlying technology. 
As people's trust plays a significant role in the growth of the cryptocurrency market, the sentiment of the general population has a substantial impact on the future market capitalization of cryptocurrencies \cite{Prajapati2020}. 
The effect of social media on crypto discourse is growing \cite{Passalis2022}. 
The sentiment expressed on social media platforms can greatly influence cryptocurrency discussions, shaping public opinion, driving market movements, disseminating information, and generating buzz. 
In 2019, over 300 million active monthly users shared their emotions in multiple languages on various social media platforms, including Twitter, Facebook, and YouTube. 
Among these platforms, Twitter has emerged as one of the most influential social media networks \cite{Matalon2021, Shanmugavadivel2022}. 
Twitter stands out as a unique platform due to its straightforward way of gauging individuals' sentiments toward a text. 
Performing real-time data analysis is also possible on the Twitter platform \cite{Binder2019}. 

Sentiment analysis is a sub-research area of computational Natural Language Processing (NLP) studies and a widely used contextual mining technique for extracting valuable and subjective information from text-based data \cite{Sasmaz2021, Tabinda2022}. 
There are several approaches to sentiment analysis, which is the process of determining the sentiment or emotion expressed in the text based on the type and size of the corpus. 
The rule-based approaches use a set of predefined rules, such as regular expressions and dictionaries, to classify text as positive, negative, or neutral \cite{Tabinda2022}. 
Machine learning approaches try to find patterns from the provided data. 
This approach involves training a machine learning model on a labeled dataset to classify new text. 
These machine-learning techniques are further expanded over supervised and unsupervised. 
Wherein supervised approaches are trained on an annotated dataset, unsupervised does not require work on experience to improve accuracy \cite{Hernández2019}. 
However, pre-trained models can be Fine-tuned on specific datasets, reducing the need for labeled data and computational resources \cite{Sharma2022}.

In text mining, sentence-level sentiment analysis has emerged as a burgeoning area of research. 
This analytical process encompasses six fundamental stages: data collection, data preprocessing, feature extraction, model training, model evaluation, and model deployment. 
Feature extraction, in particular, holds significant importance in optimizing the model's efficacy. Conventionally, Bag-of-Words (BoW) \cite{Jianqiang2018} and N-gram \cite{Chauhan2020} techniques are widely employed for this purpose. 
However, their one-hot word representation approach creates high-dimensional feature spaces and scalability challenges, failing to capture word sequences and their syntactic and semantic nuances. 
Word embedding models like word2vec \cite{Kamiş2019} and Glove \cite{Pennington2014} have gained popularity for their ability to capture word semantics in high-dimensional spaces, although they require substantial training data and are susceptible to data sparsity issues for rare or out-of-vocabulary (OVV) \cite{Tabinda2022} words, which can impact performance.

To address the abovementioned limitations, a transformer-based NLP model was proposed \cite{Vaswani2017} in the paper "Attention is All You Need." The transformer architecture is based on self-attention, which allows the model to weigh the importance of different words in a sentence when making predictions. 
In short, it creates contextual awareness between the words of a sentence. BERT \cite{Devlin2018}, T5 \cite{Raffel2019}, GPT-3 \cite{Smith2022}, and LLaMA \cite{Touvron2023} are a few examples among all pre-trained transformer-based models for NLP tasks. 
All these models have shown remarkable performance on various NLP tasks, especially in few-shot learning. However, they are less successful in zero-short learning \cite{Wei2021}. 
This paper explores a simple but powerful method called instruction tuned to improve the performance of zero-shot learning of Large Language Models (LLMs) for cryptocurrency sentiment classification tasks. 
Moreover, this paper utilizes an In-context learning (ICL) and prompt engineering method to generate effective instructions using LLMs that allow the creation of effective instructional datasets for fine-tuning the pre-trained language models to extract the cryptocurrency sentiment from social media data.

The key contributions of this paper are summarized as follows:
\begin{itemize}
  \item \textbf{Improved Model Efficiency through fine-tuning:} The study demonstrates that supervised fine-tuning and instruction tuning significantly enhance pre-trained language models' performance on unseen tasks related to cryptocurrency sentiment analysis. 
  This research experimentally shows that the average accuracy scores significantly increased after fine-tuning. 
  The findings provide compelling evidence that fine-tuning enhances model efficiency, with an average performance gain of 40\%. 
  This contributes to the practical application of fine-tuning as a powerful tool for optimizing pre-trained language model performance. 
\end{itemize}

\begin{itemize}
  \item \textbf{Benefits of Instruction Tuning and Model Scale:} Building upon previous research, the study investigates the impact of instruction tuning on different-sized language models. 
  By analyzing FLAN-T5 models of varying scales, it was discovered that larger models benefit from instruction tuning by improving generalization to new tasks. However, for small-scale models, instruction tuning had a detrimental effect on generalization, potentially due to the complete utilization of model capacity for learning the mixture of instruction tuning tasks. 
  This observation enhances our understanding of the interplay between instruction tuning and model scale, providing valuable insights for future model development and deployment. 
\end{itemize}

\begin{itemize}
  \item \textbf{Impact of model performance under different instruction tuning setups:} The conducted experimentation aimed to compare the quality of models under different instruction tuning setups, specifically focusing on the response of the instruction-based model. 
  By introducing diverse instructions of varying lengths and complexities, the study provided insights into the model's handling of different instruction types, revealing its effectiveness in understanding and executing short and simple instructions compared to long and complex ones, thereby emphasizing the importance of considering instruction characteristics in instruction tuning setups. 
\end{itemize}

\begin{itemize}
  \item \textbf{Impact of fine-tuning Corpus Size on Model Performance:} The research explores the relationship between the size of the fine-tuning corpus and the performance of language models. 
  By varying the sample size, the study investigates the influence of data availability on model performance. 
  The findings highlight the optimal corpus size of 6000 data points for achieving the highest performance for each model and the data efficiency of MiniLM in leveraging limited data.  
\end{itemize}

The remaining paper is organized as follows: Section II provides background information and reviews of related works. 
Section III presents the comprehensive architecture of the proposed model. 
Section IV showcases the analysis and metrics of the experimental results used for evaluation. 
Finally, in Section V, the paper concludes and summarizes the findings while also discussing the potential areas for future research and research limitations.

\section{Related Work}
Recently, researchers have perceived a growing interest in leveraging sentiment analysis techniques for cryptocurrency. 
Several models are available for sentiment acquisition, each with varying precision and applicability \cite{Yue2019}. 
Hasan et al. \cite{Hasan2022} investigated public sentiment analysis by employing a machine learning algorithm, namely the Support Vector Machine. 
The researchers also utilized Chi-square for feature selection to mitigate sentence noise. 
A similar approach was used by Satrya et al. \cite{Satrya2022} to determine the polarity of the sentiment. 
Additionally, they used TF-IDF weighting to transform the data from text to numeric values. 
The methodology employed by Padmalatha et al. \cite{Padmalatha2022} is based on Naive Bayes model to analyze social media opinions. 
Prasad et al. \cite{Divesh2022} designed an ensemble classifier to classify YouTube comments based on cryptocurrency. 
They used Decision Tree, K Nearest Neighbors, Random Forest Classifier, XGBoost, and a Logistic Regression base classifier to create a stacked ensemble model. 
Sasmaz et al. \cite{Sasmaz2021} studied the feasibility of automated sentiment analysis for cryptocurrencies using the Random Forest Classifier.

The Valence Aware Dictionary for Sentiment Reasoning (VADER), a widely utilized and straightforward sentiment calculation model, is commonly employed to predict cryptocurrency prices by analyzing cryptocurrency-related news and Tweets. 
Suardi et al. \cite{Suardi2022} used VADER to investigate the predictive power of information contained in social media tweets on bitcoin market dynamics. 
The extent to which Twitter sentiment analysis can predict price fluctuations for cryptocurrencies was examined by Oikonomopoulos et al. \cite{Oikonomopoulos2022}. 
They employed VADER for sentiment analysis in their study. 
Jagini et al. \cite{Jagini2023} intend to analyze the effect of tweets on the stock price of Bitcoin. 
To calculate the associated sentiment, they also used VADER. 
Parekh et al. \cite{Parekh2022} proposed a hybrid and robust DL-Gues framework for cryptocurrency price prediction. 
They also utilized a similar VADER technique in their framework to extract the polarity from the Twitter sentiment.

Regarding the recent superior performance transformers-based model, Dwivedi et al. \cite{Dwivedi2023} utilized the BERT (Bidirectional Encoder Representation) to predict the sentiments of cryptocurrency news articles. 
Kim et al. \cite{Kim2023} introduced CBITS, a Fine-tuned version of BERT designed explicitly for cryptocurrency sentiment analysis, mainly focusing on the Korean crypto market. 
Widianto et al. \cite{Widianto2023} have created a BERT model to assess sentiment analysis on cryptocurrency and NFT by utilizing data crawling and pre-processing using Rapiminer. 
The findings of Ortu et al. \cite{Ortu2021} indicate that incorporating features derived from BERT-based emotion classification of comments on GitHub and Reddit results in a notable improvement in the predictability of Bitcoin and Ethereum's hourly and daily return direction.
Despite extensive exploration of sentiment extraction methods, including rule-based and machine learning-based approaches, they often prove inadequate in cryptocurrencies due to domain-specific jargon, slang, and sentiment ambiguity not adequately covered by general-purpose texts \cite{Kim2023}. 
Rule-based systems, although less accurate, are limited by the need for more powerful linguistic resources \cite{Tabinda2022}. 
On the other hand, machine learning-based methods tend to achieve higher accuracy but require large amounts of labeled training data and significant computational resources. 
Existing literature reveals a notable gap in utilizing large language models for sentiment extraction. 
Additionally, there is a lack of research using fine-tuning techniques, despite their potential to enhance performance and adaptability. 

This study proposes integrating large language models and utilizing fine-tuning techniques in sentiment extraction to address these research gaps. 
By leveraging the capabilities of large language models and optimizing their performance through fine-tuning, this research aims to overcome the limitations of current approaches and advance the accuracy and applicability of sentiment extraction, especially in cryptocurrency. 
This contribution expects to facilitate the development of more effective and comprehensive sentiment analysis methodologies, ultimately enhancing decision-making processes in related industries and domains.

\section{Proposed System}
This section illustrates and describes the entire procedure adopted in this paper. 
It is subdivided into five sub-sections. 
Subsection III-A summarizes the proposed model along with its visualization. 
The overview of the core concept related to the proposed model is discussed in the following subsections.
\subsection{Problem Formulation}
Let $X$ be the set of tweets related to cryptocurrency where each tweet $x \in X$ is represented as a feature matrix $x \in \mathbb{R}^{1 \times n}$ thus, $X$ can be represented as
\begin{equation}
    X \in \mathbb{R}^{m \times n}
\end{equation}
where $m$ is the number of tweets and $n$ is the number of features used to represent each tweet. 
Let $Y$ be a set of sentiments represented as a target matrix 
\begin{equation}
    Y \in \mathbb{R}^{m \times 1}
\end{equation}
where $m$ is the number of tweet labels. For this experiment, only \textit{positive, negative} sentiment tweets are considered. 
Thus dataset $D$ can be represented as
\begin{equation}
    D = (X^{(i)}, Y^{(i)}|X^{(i)} \in \mathbb{R}^{(1 \times n)}, Y^{(i)} \in \mathbb{R}^{(1 \times 1)})
\end{equation}
which consisting of pairs $(X^{(i)}, Y^{(i)})$.
We can then define the model 
\begin{equation}
    f: \mathbb{R}^{(m \times n)} \rightarrow \mathbb{R}^{(m \times 1)}
\end{equation}
that maps each tweet matrix $X \in \mathbb{R}^{(m \times n)}$ to its sentiment matrix $Y \in \mathbb{R}^{(m \times 1)}$. 
The goal is to train the model $f$ on the data set $D$ such that
\begin{equation}
    \forall x \in X: f(x, \Theta) \in Y
\end{equation}
where $\Theta$ is the model parameters. 
We can use various mathematical techniques, such as gradient descent, backpropagation, and cross-entropy loss, to train/fine-tune the model $f$ on $D$. 
Moreover, We can use various optimization algorithms, such as stochastic gradient descent or Adam, to find the optimal set of parameters $\Theta^*$ that minimizes the cross-entropy loss.
The loss function can be defined as:
\begin{equation}
\begin{aligned}
   L(\Theta) = -\sum_{i=1}^m y^{(i)} log(f(x^{(i)}, \Theta)) + \\ (1-y^{(i)}) log(1 - f(x^{(i)}, \Theta))
   \end{aligned}
\end{equation}
where $m$ is the number of training examples, $x^{(i)}$ is the $i^{th}$ tweet vector in the training set, $y^{(i)}$ is the corresponding sentiment label (either 0 or 1 for negative or positive sentiment, respectively), and $f(x^{(i)}, \Theta)$ is the predicted sentiment score for $x^{(i)}$ given the current set of parameters $\Theta$. We aim to minimize the cross-entropy loss $L(\Theta)$ by finding the optimal set of parameters $\Theta^*$ that maximizes the likelihood of the training data $D$ given the model.

Once the model is trained/fine-tuned, we can use it to predict the sentiment of new tweets related to cryptocurrency by feeding their feature matrix representations into the model and obtaining the predicted sentiment matrix as output.
Mathematically the final model can be represented as 
\begin{equation}
    M_\text{eval} = (f_\text{tuned},D_\text{test})
\end{equation}
where $f_\text{tuned}$ is the trained/Fine-tuned model and $D_\text{test}$ is the evaluation dataset.

\subsection{Proposed System Architecture}
\begin{figure*}[!ht]
  \centering
  \includegraphics[width=\textwidth]{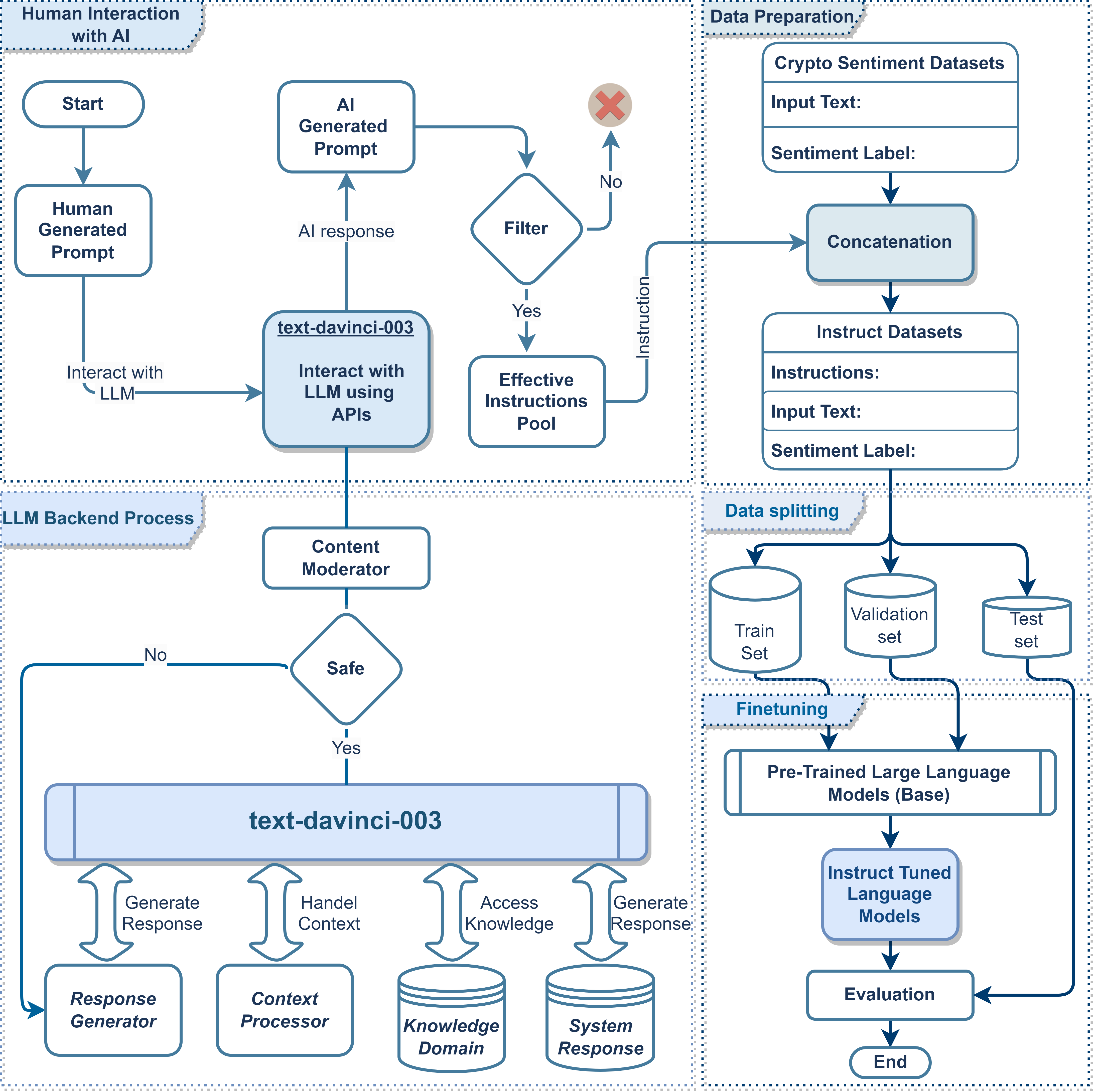}
  \caption{The Proposed System Model.}
  \label{fig: The Proposed System Model}
\end{figure*}

The design overview of our proposed model is shown in Figure \ref{fig: The Proposed System Model}. 
The proposed architecture begins with the initial user interaction step and prompt generation. 
The user engages with the OpenAI "text-davinci-003" model, initiating the prompt generation process. 
The users provide contextual information using prompt templates, which input the subsequent stages. 
The provided text is transmitted to the language model's backend through an API call, facilitating communication between the users and the large language mode. Next, a content moderator component is employed to evaluate the content. 
If the content is determined to be unsafe, the model responds with a default message. 
In the case of safe content, the system proceeds through a safety gateway to engage with the core parts of the large language model. 
This model encompasses several key components, like a response generator, context processor, knowledge domain, and system response generator. 
The system leverages its knowledge domain through interaction with these components to generate an effective response based on the provided input. 
The resulting response, an AI-generated prompt, is then delivered to the user. Upon receiving the AI-generated prompt, a filtering process takes place. 
Human feedback is vital in determining whether the prompt should progress to the subsequent stages.

The primary objective of this filtering is to generate adequate instructions by harnessing the advantages of in-context learning inherent in large language models and to prune low-quality and repeated instructions before adding them to the task pool. 
Finally, the prompt generation is refined by incorporating user feedback and capitalizing on the model's contextual understanding to enhance its efficacy. 
The process involving user interaction, prompt generation, filtering, and human feedback collectively constitutes the in-context learning and prompt engineering phase.
The overall process of generating instructions using in-context learning can be seen in Algorithm \ref{Algo_enerating instructions using in-context learning}.
This phase aims to iteratively improve the prompt generation procedure by leveraging user feedback and maximizing the model's contextual comprehension capabilities.

Following the generation of effective instructions, the subsequent objective is to create an augmented dataset. 
This dataset is constructed by concatenating the introductions with the original dataset, ensuring a comprehensive collection of relevant information for training purposes. 
Algorithm \ref{Algo_generating augmented crypto sentiment dataset} shows the process of generating an augmented crypto sentiment dataset.
The instructional dataset is then divided into separate training and validation sets. 
Three additional datasets are withheld to measure zero-shot performance, remaining untouched for evaluation. 
The large language models undergo fine-tuning at this stage, utilizing the instructional dataset. 
The initial model weights are modified, resulting in a newly instructed, Fine-tuned model version. 
The overall fine-tuning process of a large language model can bee observed in Algorithm \ref{Algo_fine-tuning Algorithm}.
After the fine-tuning process, the performance of various models is evaluated. 
This evaluation provides valuable insights into the efficacy and effectiveness of the in-context learning and prompt engineering techniques employed in instructing fine-tuning for Zero-shot learning. 

\begin{algorithm}
    \caption{Generating instructions using in-context learning.}\label{alg:gen_inst}
    \hspace*{\algorithmicindent} \textbf{Constant} model, model, temp, max\_len, top\_p, penalty\\
    \hspace*{\algorithmicindent} \textbf{Input} A human-generated prompt $p \in P$.\\
    \hspace*{\algorithmicindent} \textbf{Output} An effective instructions pool $Y$.\\
    \begin{algorithmic}[1]
        \STATE Define the Language Model (LLM) as a function $G: P \rightarrow R$ that maps a given prompt $p \in P$ to a model-generated response $r \in R$.
        \STATE $r \gets G(p; mode, model, temp, max\_len, top\_p)$
        \STATE Pass the response through  a filter $F: R \rightarrow (True, False)$
        \FOR{response $r \in R$}
            \IF{$F(R)=True$}
                \STATE Add $r$ to the effective instructions pool $Y$.
            \ELSE
                \STATE Discard the response $r$.
            \ENDIF
        \ENDFOR
        \STATE \textbf{Return} Effective instructions pool $Y$
    \end{algorithmic}
\label{Algo_enerating instructions using in-context learning}
\end{algorithm}

\begin{algorithm}
    \caption{Generating augmented crypto sentiment dataset.}\label{alg:gen_aug}
    \hspace*{\algorithmicindent} \textbf{Input} Effective instructions pool $Y$, and Crypto sentiment dataset $X$.\\
    \hspace*{\algorithmicindent} \textbf{Output} Augmented crypto sentiment dataset $X_{augmented}$.\\
    \begin{algorithmic}[1]
        \STATE Let function $C: x \rightarrow x$ which cleans a dataset entry.
        \STATE $X_{filtered} \subseteq X $which contains only non-neutral sentiments.
        \FOR{$i \gets (1 \dots len(X_{filtered}))$}
            \STATE $X_{cleaned}^{(i)} \gets C(X_{filtered}^{(i)})$.
        \ENDFOR
        \STATE Select an instruction $y \in Y$.
        \STATE Let function $A: (x, y) \rightarrow x$ which augments each entry $x \in X$ with an instruction $y$.
        \FOR{$i \gets (1 \dots len(X_{cleaned}))$}
            \STATE $X_{augmented}^{(i)} \gets A(X_{cleaned}^{(i)}, y)$.
        \ENDFOR
        \STATE \textbf{Return} Augmented dataset $X_{augmented}$
    \end{algorithmic}
\label{Algo_generating augmented crypto sentiment dataset}
\end{algorithm}

\begin{algorithm}
    \caption{Fine-tuning of a pre-trained language model.}\label{alg:cap}
    \hspace*{\algorithmicindent} \textbf{Constant} random\_seed, input sequence size, number of layers, number of hidden layer nodes, number of classifier outputs.\\
    \hspace*{\algorithmicindent} \textbf{Input} Training set $(X_{train}^{(1)}, Y_{train}^{(1)}), … , (X_{train}^{(M)}, Y_{train}^{(M)})$.\\
    \hspace*{\algorithmicindent} \textbf{Output} Trained language model network parameters..\\
    \begin{algorithmic}[1]
        \FOR{$epoch \in epochs$}
            \FOR{$batch \in batches$}
                \FOR{$i \gets (1 \dots len(batch))$}
                    \STATE $M_{train}^{(i)} \gets tokenizer(X_{train}^{(i)})$
                    \STATE $\hat{M}_{train}^{(i)} \gets model(M_{train}^{(i)})$
                \ENDFOR
                \STATE $loss \gets E(Y_{train}^{(batch)}, Y_{train}^{(batch)})$
                \STATE Calculate $\nabla \Theta$ for backpropagation.
                \STATE Adjust parameters using an optimizer to minimize the loss.
            \ENDFOR
        \ENDFOR
        \STATE \textbf{Return} Task-specific Fine-tuned model.
    \end{algorithmic}
\label{Algo_fine-tuning Algorithm}
\end{algorithm}

\subsection{Pre-trained Language Models (PLMs)}
Recently, there has been a significant emphasis on pre-trained language models (PLMs) that utilize self-supervised learning on extensive raw text data \cite{Qiu2020}. 
Notable examples of such models include GPT-3 \cite{Smith2022}, PaLM \cite{Chowdhery2022}, Chinchilla \cite{Hoffmann2022}, LLaMA \cite{Touvron2023}, and Falcon 40B \cite{Penedo2023}. 
By training on large-scale texts using self-learning tasks like masked word prediction, sentence sequence recognition, text completion, and text generation \cite{Vaswani2017, Du2023} PLMs acquire a comprehensive understanding of language. 
In addition, these models enhance the semantic representation of words by considering contextual dynamics and provide a unified framework for various NLP tasks. 
Currently, there are three standard models \cite{Wu2023} structures in PLMs: autoregressive language models, autoencoding language models, and hybrid language models. 
Representative models for each design are GPT \cite{Smith2022}, BERT \cite{Devlin2018}, and T5 \cite{Raffel2019}, respectively. 
Autoregressive language models follow a standard approach where language modeling is done decoder-only, predicting words one by one through one-way language encoding-decoding and token-by-token prediction of words. 
Autoencoding language models randomly mask words in a sentence, use bidirectional encoding to capture context, and then predict the masked words based on the encoded information. 
Finally, hybrid language models combine the approaches of the previous two models. 
They mask words randomly in a sentence, apply bidirectional encoding, and predict subsequent words step by step by inputting the earlier text in one direction \cite{Wu2023}.

The advancement of artificial intelligence technology has demonstrated that LLMs can acquire a deeper understanding of language and exhibit stronger capabilities in understanding and generating data. 
These models learn abstract knowledge from raw data, resulting in better generality and generalization. 
The autoregressive language model adopted by GPT-3 and its subsequent models, such as GPT3.5 and GPT4, has proven advantageous in utilizing natural language for various tasks in different fields \cite{Wu2023}. 
Initially, it was believed that increasing the number of parameters in models would lead to better performance. 
However, recent research by Hoffmann et al. \cite{Touvron2023} has shown that smaller models trained on more data can achieve the best performances given a specific computing budget. 

\begin{figure}[ht]
    \centering
    \includegraphics[width=0.43\textwidth]{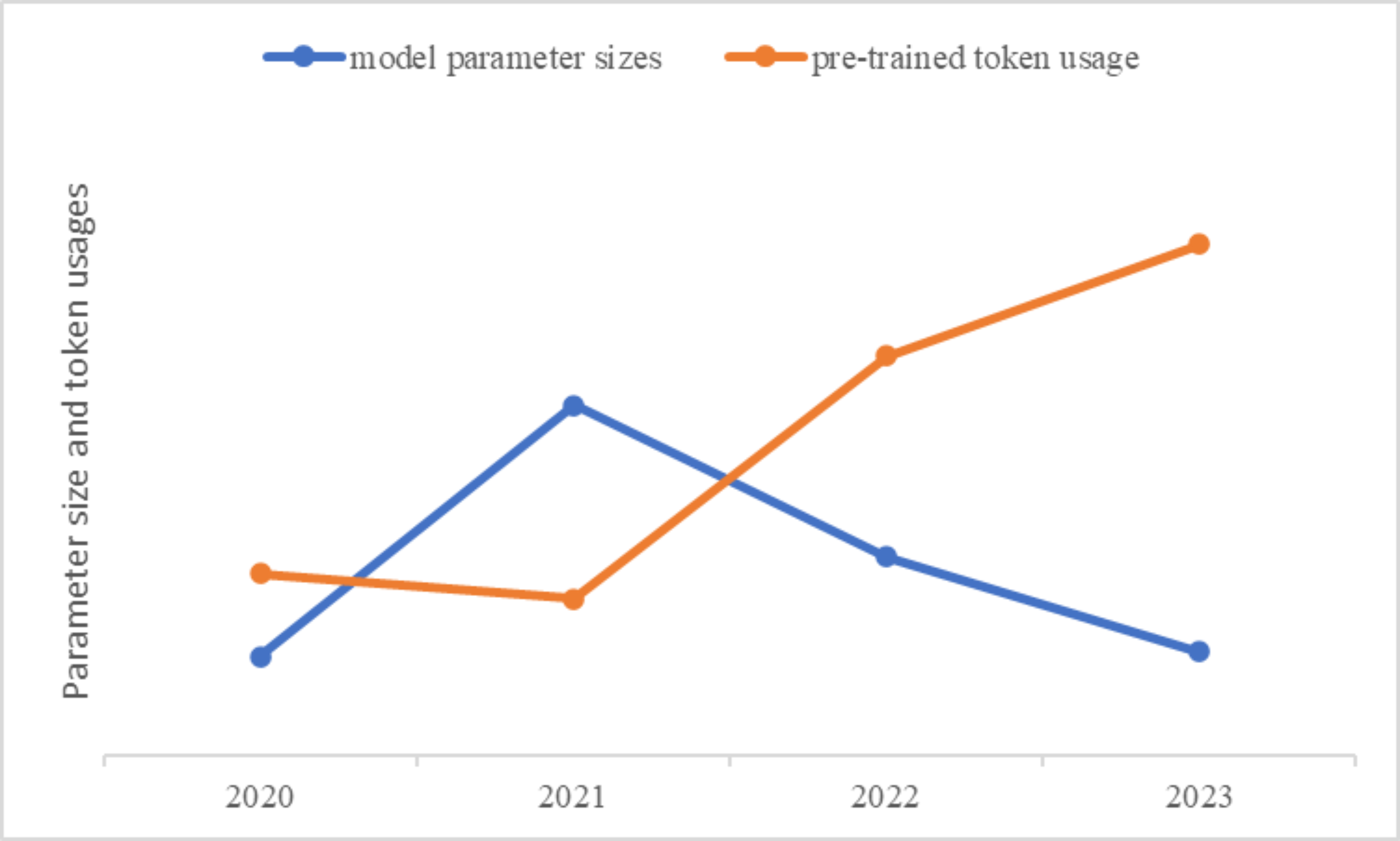}
    \caption{Yearly trend analysis of model parameter size and token usage.}
    \label{fig: Yearly trend analysis based on model parameter size and token usages}
\end{figure}

According to Figure \ref{fig: Yearly trend analysis based on model parameter size and token usages}, the data analysis shows a clear trend of increasing pre-trained token usage over the years. 
In the initial years (2020-2021), models exhibited relatively low token counts during the pre-training phase. 
However, in recent years (2022-2023), there has been a notable surge in the number of pre-trained tokens utilized by language models. 
Such expansion signifies model developers' recognition of the benefits of leveraging more extensive and diverse pre-training corpora, enabling improved contextual understanding and enhanced performance in downstream tasks. 

Contrary to the trend observed in pre-trained token usage, model parameter sizes exhibit a different pattern.
In 2021 and 2022, models with considerably large parameter sizes emerged. 
However, in 2023, a noticeable decrease in model parameter sizes is observed. 
This shift suggests a growing focus on optimizing computational efficiency and addressing the resource-intensive nature of large models. 
As a result, model developers are actively exploring methods to achieve comparable performance with fewer parameters, which may reduce computational costs and carbon footprint. 

\begin{figure}[ht]
    \centering
    \includegraphics[width=0.43\textwidth]{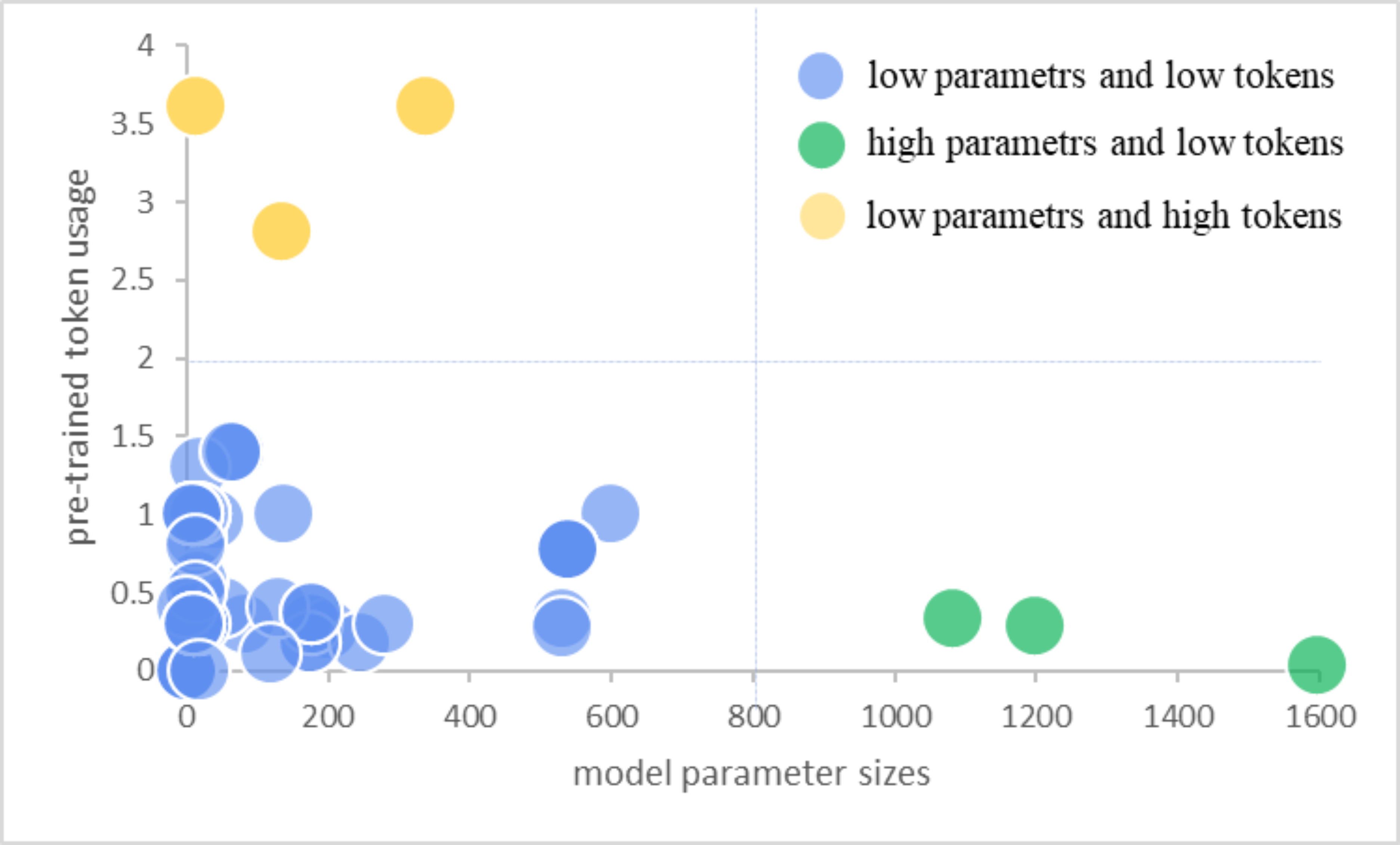}
    \caption{Cluster distribution analysis considering model parameters and token utilization.}
    \label{fig: Cluster distribution based on model parameter size and token usages}
\end{figure}

According to Figure \ref{fig: Cluster distribution based on model parameter size and token usages}, language models can categorize into three distinct clusters: Models with low token usage and small parameter sizes, representing the majority, are suitable for resource-constrained environments. 
Models with substantial token counts but relatively small parameter sizes exhibit an exciting trade-off, leveraging massive amounts of pre-training data while keeping the parameter sizes reasonably small. 
Models with low token usage and large parameter sizes prioritize performance and strike a balance between computational resources and model capacity.
The observed trends in language model development hold significant implications for researchers and practitioners. 
The increasing usage of pre-trained tokens emphasizes the importance of diverse and extensive training data for capturing the degree of language understanding. 
Conversely, the fluctuation in model parameter sizes highlights ongoing efforts to balance model capacity and computational efficiency. 
Further research is required to explore novel techniques and architectures that optimize the trade-offs in token usage and parameter sizes.

\subsection{fine-tuning of Large Language Models (LLMS)}
LLMs have exhibited exceptional abilities in various NLP tasks \cite{Brown2020, Chowdhery2022, Zhang2022}. 
Nevertheless, these models can sometimes display unintended behaviors, such as generating false information, pursuing inaccurate objectives, and producing harmful, misleading, and biased expressions \cite{Ouyang2022, Zhao2023}. 
In the pre-training stage, pre-trained language models acquire non-task-specific language knowledge. 
The subsequent fine-tuning stage facilitates task-specific adjustments of the model, enabling it to be utilized for various downstream tasks \cite{Wei2023}. There are two primary approaches \cite{raschka2023fine} to adapt the pre-trained language models for target tasks: feature-based approach and fine-tuning. 
The feature-based approach involves loading a pre-trained LLM and utilizing it on a target dataset. The primary focus is creating output embeddings for the training set, which can be used as input features for a classification model. 
While this approach is often used for embedding-centric models like BERT, embeddings can also be extracted from generative GPT-style models like "text-embedding-ada-002". 
The classification model can be any desired model, such as a logistic regression model, random forest, or XGBoost. However, linear classifiers, specifically logistic regression, have demonstrated superior performance \cite{raschka2023fine}.

fine-tuning is essential for adapting pre-trained language models to perform specific tasks using labeled training data. 
Initially, a pre-trained language model is used as a starting point and then Fine-tuned on a task-specific dataset with labeled examples. 
This process is called supervised fine-tuning (SFT). 
This SFT process is necessary to apply PLMs to tasks like sentence classification, named entity recognition, and question-answering. 
Unlike pre-training, this fine-tuning requires less data, typically around 100k words \cite{Shah2022pre}. 
During SFT, a task-specific layer is added to the PLMs, and the model parameters, including those of the task-specific layer, are updated through gradient descent using an appropriate loss function \cite{raschka2023fine}. 
One advantage of the PLMs is the ability to freeze specific layers while fine-tuning the remaining ones, potentially improving performance. 
However, it has been observed that freezing too many layers may lead to poor performance \cite{Shah2022pre}. 
The power and popularity of PLMs come from the fact that the pre-training process only needs to be done once in a task-agnostic manner. 
Subsequently, a simple and more cost-effective fine-tuning process is sufficient for each specific task. 
This is possible because the dataset size required for fine-tuning is considerably smaller, reducing time and resource requirements \cite{Thangarasa2023}. 

Another form of fine-tuning is instruction tuning (IT) – fine-tuning language models on a collection of datasets described via instructions. Fundamentally, instruction tuning involves fine-tuning pre-trained LLMs using a set of formatted instances in natural language form \cite{Wei2021}. 
This approach is closely related to SFT \cite{Ouyang2022}. 
The initial step involves gathering or constructing instances formatted as instructions to initiate instruction tuning. 
Subsequently, these formatted instances are employed to finetune LLMs using supervised learning techniques, such as training with the sequence-to-sequence loss. 
Following instruction tuning, LLMs exhibit enhanced generalization capabilities towards unseen tasks \cite{Wei2021,Sanh2021, Chung2022}, even within a multilingual context \cite{Muennighoff2022}. 
IT is a more efficient alternative to pre-training, as it relies on a moderate number of instances for training. 
This supervised training process introduces distinct optimization considerations compared to pre-training, including utilizing a sequence-to-sequence loss as the training objective and special attention to factors like smaller batch size and learning rate in the optimization configuration \cite{Chung2022}. 
IT significantly impacts LLMs by improving performance across various models and enhancing task generalization by enabling LLMs to understand and follow natural language instructions \cite{Zhao2023, Workshop2022}. 

\subsection{In-Context Learning (ICL)}
In-Context Learning (ICL) pertains to comprehending the context of the input information and leveraging it accurately to produce the intended output. It exhibits similarities to the human decision-making process, wherein individuals learn from analogy \cite{Dong2022}. 
In LLM utilization, ICL was initially introduced as a distinctive prompting method, specifically alongside GPT-3 \cite{Brown2020}, and has since emerged as a prominent approach \cite{Wu2023}. 
In contrast to supervised learning, which involves a training phase that utilizes backward gradients to modify model parameters, ICL does not engage in parameter updates but instead performs direct predictions on pre-trained language models. 
The effectiveness of ICL lies in its dependence on the existing knowledge of the model and its ability to decipher the concealed pattern present in the demonstrations, thus facilitating precise predictions. 
The approach holds significant potential in situations that demand swift adaptation to new tasks, as it prevents the need for extensive training periods \cite{layton2023prompt}. 
The vanilla GPT-3 model shows considerable potential for ICL, as pre-training adaptation has been shown to enhance its capabilities \cite{Montgomery1980}. 
Moreover, the efficacy of ICL is contingent upon various parameters such as the choice of prompting template, in-context examples, and their sequence \cite{Min2021}. 
Despite being plausible, the underlying mechanism of ICL remains obscure, and only a few studies have provided initial insights into its workings \cite{Zhang2022, Dai2022}. 

The context window size in LLMs \cite{Wu2023}, such as GPT-4, determines the maximum amount of input text the model can consider when generating its output. With the release of GPT-4, the context window's size doubled. GPT-3 was limited to 2048 tokens. 
The context window for the GPT-4 API is 8195 tokens, and a 32K context window exists in the most significant model. 
However, its performance typically falls short of fine-tuning, as it needs to update the model's parameters for a specific task, which may limit its adaptability to task-specific nuances \cite{raschka2023fine}. 
Moreover, there are potential risks associated with ICL. The risk of prejudice and misinformation is a significant concern, as the LLMs cannot fact-check the input provided as part of the prompt, resulting in the possibility of incorporating erroneous and biased information into any generated output, including fabricated news or blog posts \cite{Lopez-Lira2021}.

\subsection{Prompt Engineering}
As technological progress continues to unfold, the significance of prompt engineering grows in parallel. 
The ascendancy of large language models, such as ChatGPT \cite{Wu2023}, necessitates a skillset that can proficiently engage with them. 
By furnishing appropriate prompts, adherence to prescribed norms and the automation of processes can be ensured, resulting in outputs that align with desired quality and quantity benchmarks. 
Like programming, prompts facilitate the customization of interactions with large language models, optimizing their utilization to meet specific requirements \cite{White2023}.

\begin{figure}[ht]
    \centering
    \includegraphics[width=0.42\textwidth]{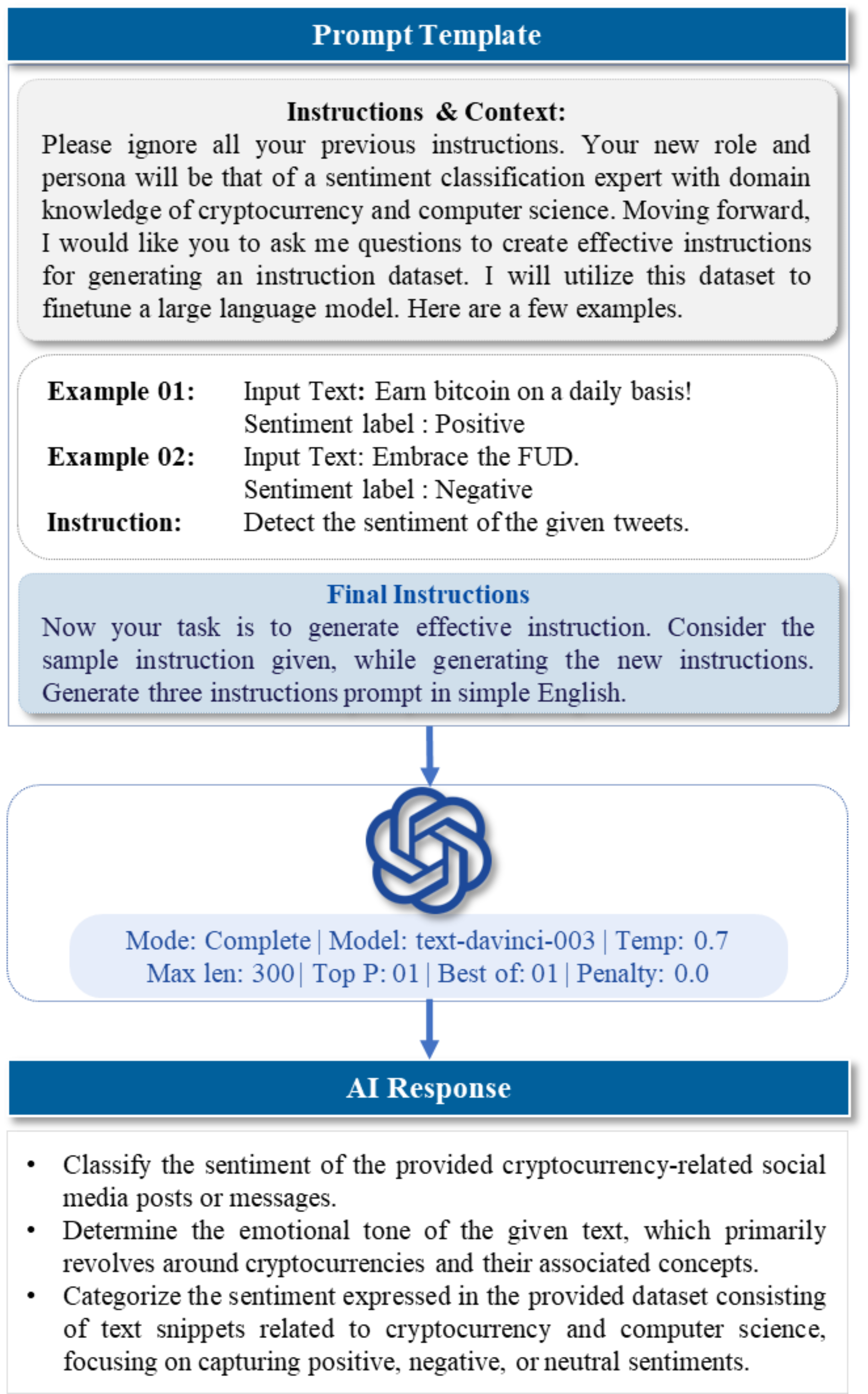}
    \caption{Prompt engineering template.}
    \label{fig: Prompt engineering template}
\end{figure}

A prompt comprises a specific set of directives furnished to a LLM, enabling customization and refinement of the model's capabilities through programming \cite{Liu2021}. 
In short, it is a text that provides context and instructions for LLMs to generate a response. Prompts are pivotal in facilitating LLMs to undertake extensive linguistic tasks, including language translation, text classification, text summarization, question answering, and even producing human-like and coherent text \cite{Lopez-Lira2021}. 
The use of prompting techniques in NLP tasks has been extensively studied, covering both zero-short and few-shot settings, as indicated by various research papers \cite{Gao2021, Le2021, Schick2020, Strobelt2023}. 
Current prompt-based models predominantly rely on Transformers \cite{Vaswani2017}. 
Prompting is particularly prevalent in online demonstrations, where generative models are interacted with using Transformers as assistive agents. 
By leveraging prompt engineering, the challenging task of language understanding can be addressed with improved efficiency \cite{Maddigan2023}. 
The 'show-and-tell' technique \cite{openai_best}, where examples and instructions are provided within the prompt, is the most effective approach for obtaining the desired output from the LLMs. 
We provide an example prompt in our LLM to demonstrate how prompts for LLMs can be engineered to elicit desired outcomes. 
The structure of the prompt is presented in Figure \ref{fig: Prompt engineering template}. 
Prompt engineering possesses several notable advantages: Firstly, it enables the pre-training of the language model on extensive volumes of raw text. Secondly, by defining a new prompting function, the model demonstrates the capability for few-shot or even zero-shot learning, effectively adapting to new scenarios with limited or no labeled data \cite{Liu2021}.

\section{Result Analysis}
This result analysis section is subdivided into three sub-sections. Together, they describe the dataset, narrate the evaluation metrics, illustrate the detailed experimental result, and compare the performance of the models. The experiments were repeated multiple times to check for anomalies and to get rid of any bias.

\subsection{Dataset Description and Pre-processing}
Four datasets, namely the Neo, Reddit, Bitcoin sentiment, and Cryptocurrency sentiment datasets are utilized in the experimental analysis. 
These datasets are the foundation for conducting comprehensive investigations and analyses related to sentiment analysis in cryptocurrency. Table \ref{table_dataset_description} shows the statistics. 

The Neo dataset consists of tweets related to cryptocurrencies, specifically focused on the sentiment towards the Neo cryptocurrency. 
The dataset contains 12,000 tweets, evenly distributed between positive and negative emotions, with 6,000 tweets in each category. 
The Reddit dataset contains posts and comments extracted from the popular social media platform Reddit. 
The dataset focuses on the sentiment expressed towards various cryptocurrencies. 
It includes a total of 562 posts. Among these posts, 302 are classified as positive sentiment, while 260 are classified as negative sentiment. 
The Bitcoin sentiment dataset consists of tweets specifically related to the Bitcoin cryptocurrency. 
The dataset comprises 1,029 tweets, with 779 tweets classified as positive sentiment and 250 tweets classified as negative sentiment. 
It is important to note that this dataset exhibits an imbalance in sentiment distribution, with more positive than negative tweets taken intentionally for the experiments.
The Cryptocurrency sentiment dataset is a collection of tweets that cover a wide range of cryptocurrencies. 
The dataset includes 500 tweets, evenly split between positive and negative sentiments, with 250 tweets in each category.
\begin{table}[!ht]
    \centering
    \caption{Volume of all tweets and volume of tweets for each sentiment label and dataset.}
    \begin{tabular}{lcc}
    \hline
    Description    & \multicolumn{1}{l}{Volume} & \multicolumn{1}{l}{Percentage} \\ \hline
    All tweets     & 14,091                     & 100.00\%                       \\
    Positive label & 7,331                      & 52.03\%                        \\
    Negative label & 6,760                      & 47.97\%   \\ 
    Neo dataset & 12,000 & 85.16\% \\
    Bitcoin sentiment dataset & 1,029 & 7.30\% \\
    Reddit dataset & 562 & 3.99\% \\
    Cryptocurrency sentiment dataset & 500 & 3.55\% \\
    \hline
    \end{tabular}
\label{table_dataset_description}
\end{table}

This study uses the Neo dataset to finetune the pre-trained language models. 
The other datasets, Reddit, Bitcoin sentiment, and Cryptocurrency sentiment datasets, are employed to evaluate the performance of the Fine-tuned models on unseen tasks. 
The concept of unseen tasks is defined based on prior work, which disallows the same dataset to appear during training. 
By separating the training and evaluation datasets in this manner, the study aims to provide a robust assessment of the model's ability to generalize and perform effectively on new, unseen data.

\subsection{Evaluation Metrics}
To evaluate the effectiveness of the Fine-tuned models, accuracy (binary classification accuracy) and F1-score were selected as the evaluation metrics. The calculation of binary classification accuracy can be represented using the following equation:

\begin{equation}
    \mathcal{T}\ = \frac{\theta + \emptyset }{\theta + \Lambda + \emptyset + \Psi}
\end{equation}
where \( \mathcal{T} \) represents the Accuracy (the percentage of the correct number predicted by the model to the total number of samples).
\( \theta \) corresponds to the true positive (the predicted result is positive, and the true result is also positive).
\( \emptyset \) is the true negative (the predicted result is negative, and the true result is also negative).
\( \Lambda \) corresponds to the false positive (the predicted result is positive, but the true result is negative).
\( \Psi \) represents the true false negative (the predicted result is negative, but the true result is positive).

For evaluating the value of F1 score, which is the harmonic mean of precision and recall, the equation would be represented as 
\begin{equation}
    F1 = \frac{2}{\text{Precision } ^ {-1} + \text{Recall } ^ {-1}}
\end{equation}
\begin{equation}
    F1 = \frac{\theta}{\theta + 0.5(\emptyset + \Psi)}
\end{equation}
where the precision is defined as the number of true positives over the number of true positives plus the number of false positives, and recall is the number of true positives over the number of true positives plus the number of false negatives.

\subsection{Comparison Experiment And Discussion}
This section will describe the research questions, the experiment detail, the result, and the discussion

\subsubsection{How does supervised fine-tuning and instruction tuning impact the efficiency of pre-trained language models in terms of performance on unseen tasks?}
The research question of whether fine-tuning improves the model efficiency was investigated through a series of experiments. 
Three pre-trained language models (DistilBert, MiniLM, and FLAN-T5-Base) and four datasets were used to train and evaluate the performance of the models. 
This work was mainly focused on the zero-shot setup under the default parameter setup except for the learning rate, batch size, and no of epochs. 
The models were trained for a maximum of 3 epochs, and the Adam optimizer with a learning rate of 0.00002, and batch size of 8 was utilized. 

Based on the experimental setup the pre-trained language models were divided into three groups: untuned model, SFT model, and IT model. 
Untuned model is the vanilla LM having the original checkpoint (only pre-training, no additional fine-tuning). 
The SFT model is considered the fine-tuning setups in a standard supervised way without instruction. 
It follows a no-template setup, only inputs and outputs were given to the model (e.g., for text classification, the input would be "Earn bitcoin on a daily basis!" and the output would be "Positive"). 
On the other hand, the IT model is considered fine-tuning setups in a standard supervised way with natural instruction. 
For IT model, first the instruction and instance input were concatenated to make a single prompt (e.g., for text classification. the input would be "Detect the sentiment of the given text, Text: Earn bitcoin on a daily basis!" and the output would be "Positive") and then trained the model to generate the instance output. 

Tables \ref{table_classication results on bitcoin dataset},  \ref{table_classication results on reddit dataset}, and  \ref{table_classication results on cryptocurrency dataset} present the results of experiments that were conducted to classify cryptocurrency-related tweets into two categories: positive and negative. The performances were compared between three language models with their vanilla, supervised Fine-tuned, and instruction-based Fine-tuned versions in terms of accuracy, F1 score, precision, and recall.

\begin{figure}[ht]
  \centering
  \includegraphics[width=.45\textwidth]{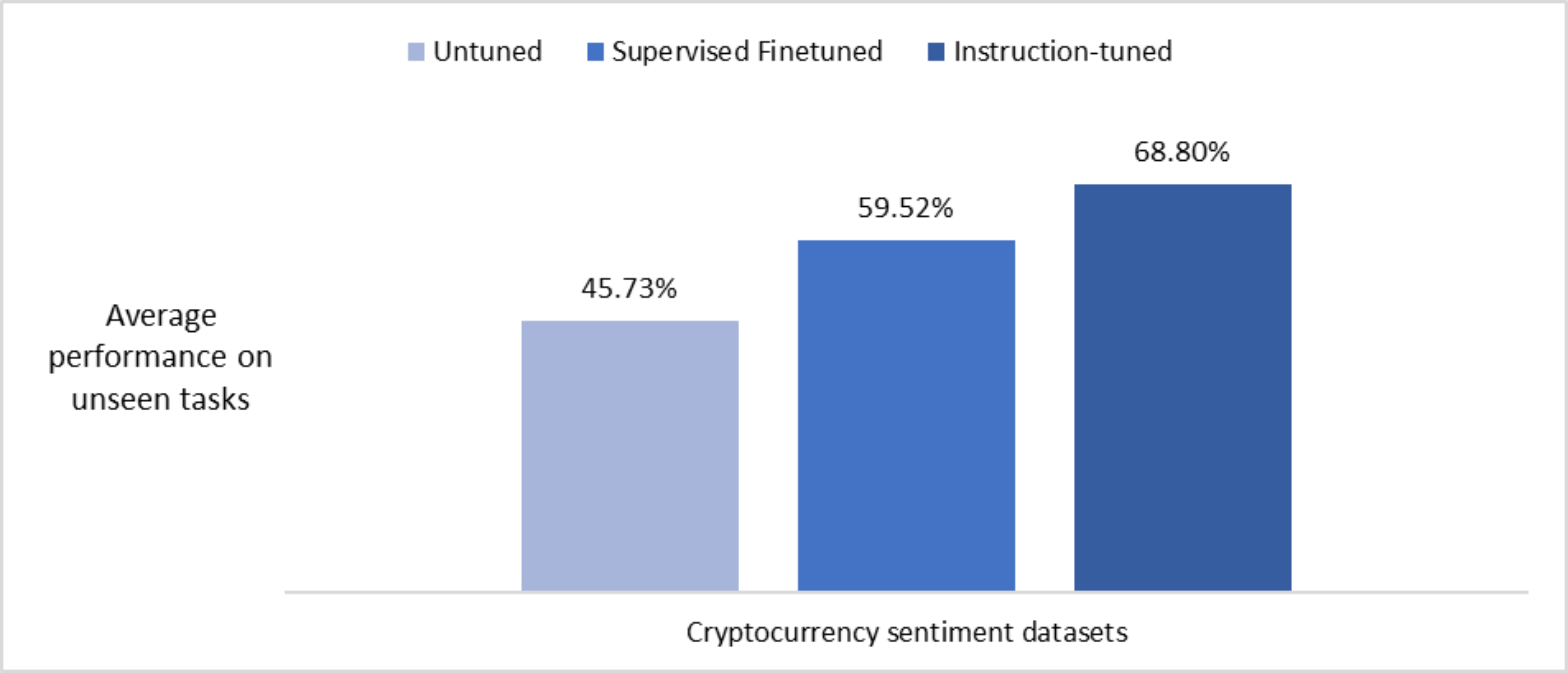}
  \caption{Zero-shot performance analysis between untuned, supervised, and instruction-base Fine-tuned models on unseen tasks.}
  \label{fig: Performance analysis of untuned and Fine-tuned models}
\end{figure}

The overall analysis of the results based on the accuracy score is shown in Figure \ref{fig: Performance analysis of untuned and Fine-tuned models}, which indicates that fine-tuning improves the model performance dramatically. 
The baseline performance for this study was established with the vanilla LM models. The DistiBERT-vanilla model achieved an accuracy score of 35.66\%, 58.95\%, and 60.33\% across three datasets Bitcoin sentiment dataset, Reddit dataset, and Cryptocurrency sentiment dataset respectively. The MiniLM-vanilla model achieved an accuracy score of 41.67\%, 46.67\%, and 51.25\%. And the Flan-T5-Base-vanilla model achieved an accuracy score of 25.10\%, 46.07\%, and 45.89\%. The average accuracy score of all three vanilla LM models across all three different datasets is 45.73\%. 
However, after fine-tuning, the average performance improved significantly to 59.52\% for SFT model and 68.80\% for IT model. 
The experimental results provide compelling evidence that fine-tuning enhances model efficiency. 
The substantial improvements in accuracy scores demonstrate the effectiveness of fine-tuning across different datasets and models. 
The average performance gain of 40\% highlights the significant impact of fine-tuning on model performance, indicating its potential for practical applications.

Additionally, a comparison was made between the SFT and IT models to identify the role of instructions. 
The results revealed that the Fine-tuned models with instructions outperformed their counterparts without instructions. 
On average, the performance of the IT model was 16\% higher than the SFT model. 
The consistent improvements observed across multiple models and datasets reinforce the generalizability of the findings. 
The results indicate that fine-tuning can be a valuable technique for optimizing model efficiency in various domains and tasks, providing researchers and practitioners with a powerful tool to enhance model performance.
\begin{table}[!ht]
    \centering
    \caption{Classification results on Bitcoin sentiment dataset.}
    \begin{tabular}{lllll}
    \hline
    Model      & Accuracy & F1 score & Precision & Recall   \\ \hline
    DistilBERT-vanilla   & 35.66\%           & 24.42\%           & 62.60\%            & 15.97\%           \\ 
    MiniLM-vanilla       & 41.67\%           & 27.99\%           & 25.00\%            & 33.33\%           \\ 
    Flan-T5-Base-vanilla & 25.10\%           & 0.26\%            & 0.78\%             & 0.16\%            \\ \hline
    \textit{Avg\_vanilla}         & 34.14\%           & 17.56\%           & 29.46\%            & 16.49\%           \\ \hline
    DistilBERT-SFT       & 42.83\%           & 41.28\%           & 70.68\%            & 31.68\%           \\ 
    MiniLM-SFT           & 66.07\%           & 63.03\%           & 70.91\%            & 61.21\%           \\ 
    Flan-T5-Base-SFT     & 65.89\%           & 62.13\%           & 72.57\%            & 60.13\%           \\ \hline
    \textit{Avg\_SFT}             & 58.26\%           & 55.48\%           & 71.39\%            & 51.00\%           \\ \hline
    DistilBERT-IT        & 68.12\%           & 70.93\%           & \textbf{91.87\%}   & 61.53\%           \\ 
    MiniLM-IT            & 70.38\%           & 78.01\%           & 78.59\%            & 81.40\%           \\ 
    Flan-T5-Base-IT      & \textbf{75.00\%}  & \textbf{83.98\%}  & 75.00\%            & \textbf{100.00\%} \\ \hline
    \textit{Avg\_IT}              & 71.17\%           & 77.64\%           & 81.82\%            & 80.98\%           \\ \hline
    \textbf{Best\_Score}          & \textbf{75.00\%}           & \textbf{83.98\%}          & \textbf{91.87\%}            & \textbf{100.00\%}         \\ \hline
    \end{tabular}
\label{table_classication results on bitcoin dataset}
\end{table}

\begin{table}[!ht]
    \centering
    \caption{Classification results on Reddit dataset.}
    \begin{tabular}{lllll}
    \hline
    Model      & Accuracy & F1 score & Precision & Recall   \\ \hline
    DistilBERT-vanilla                   & 58.95\%           & 30.79\%           & 49.60\%            & 23.45\%          \\ 
    MiniLM-vanilla                       & 46.67\%           & 0.96\%            & 1.90\%             & 0.77\%           \\ 
    Flan-T5-Base-vanilla                 & 46.07\%           & 1.05\%            & 2.86\%             & 0.64\%           \\ \hline
    \textit{Avg\_vanilla}                & 50.56\%           & 10.93\%           & 18.12\%            & 8.29\%           \\ \hline
    DistilBERT-SFT                       & 66.07\%           & 63.03\%           & 70.91\%            & 61.21\%          \\ 
    MiniLM-SFT                           & 60.71\%           & 57.00\%           & 63.48\%            & 57.08\%          \\ 
    Flan-T5-Base-SFT                     & 53.75\%           & 67.42\%           & 53.70\%            & \textbf{99.71\%} \\ \hline
    \textit{Avg\_SFT}                    & 60.18\%           & 62.48\%           & 62.70\%            & 72.67\%          \\ \hline
    DistilBERT-IT                        & 64.82\%           & 55.60\%           & \textbf{76.86\%}   & 48.04\%          \\ 
    MiniLM-IT                            & 63.75\%           & 62.03\%           & 65.06\%            & 65.94\%          \\ 
    Flan-T5-Base-IT                      & \textbf{74.29\%}  & \textbf{75.00\%}  & 70.19\%            & \textbf{86.60\%} \\ \hline
    \textit{Avg\_IT}                     & 67.62\%           & 64.21\%           & 70.70\%            & 66.86\%          \\ \hline
    \textbf{Best\_Score}                 & \textbf{74.29\%}  & \textbf{75.00\%}  & \textbf{76.86\%}   & \textbf{99.71\%} \\ \hline
    \end{tabular}
\label{table_classication results on reddit dataset}    
\end{table}

\begin{table}[!ht]
    \centering
    \caption{Classification results on Cryptocurrency sentiment dataset.}
    \begin{tabular}{lllll}
    \hline
    Model      & Accuracy & F1 score & Precision & Recall   \\ \hline
    DistilBERT-vanilla                   & 60.33\%           & 30.76\%           & 49.80\%            & 23.15\%          \\ 
    MiniLM-vanilla                       & 51.25\%           & 45.64\%           & 35.83\%            & 66.67\%          \\ 
    Flan-T5-Base-vanilla                 & 45.89\%           & 0.95\%            & 2.86\%             & 0.57\%           \\ \hline
    \textit{Avg\_vanilla}                & 52.49\%           & 25.78\%           & 29.50\%            & 30.13\%          \\ \hline
    DistilBERT-SFT                       & 65.89\%           & 62.13\%           & 72.57\%            & 60.13\%          \\ 
    MiniLM-SFT                           & 60.54\%           & 58.19\%           & 67.19\%            & 57.16\%          \\ 
    Flan-T5-Base-SFT                     & 53.93\%           & 68.49\%           & 53.88\%            & \textbf{99.71\%} \\ \hline
    \textit{Avg\_SFT}                    & 60.12\%           & 62.94\%           & 64.54\%            & 72.34\%          \\ \hline
    DistilBERT-IT                        & 64.82\%           & 55.60\%           & \textbf{76.86\%}   & 48.04\%          \\ 
    MiniLM-IT                            & 63.75\%           & 62.03\%           & 65.06\%            & 65.94\%          \\ 
    Flan-T5-Base-IT                      & \textbf{74.29\%}  & \textbf{75.00\%}  & 70.19\%            & \textbf{86.60\%} \\ \hline
    \textit{Avg\_IT}                     & 67.62\%           & 64.21\%           & 70.70\%            & 66.86\%          \\ \hline
    \textbf{Best\_Score}                 & \textbf{74.29\%}  & \textbf{75.00\%}  & \textbf{76.86\%}   & \textbf{99.71\%} \\ \hline
    \end{tabular}
\label{table_classication results on cryptocurrency dataset}    
\end{table}
\subsubsection{How the benefits of instruction tuning are affected by model scale?}
Based on the study conducted by Brown et al. \cite{Brown2020}, which revealed that zero and few-shot capabilities of larger language models substantially improve for larger models, the present research delves into investigating how the scale of the model influences the benefits of instruction tuning. 
The impact of instruction tuning was evaluated across FLAN-T5 models of different sizes: small (80M), base (250M), and large (780M), based on their parameters. The model's architecture and their comparative analysis have been summarized in Table \ref{table_FLAN_architecture_detail}.

\begin{figure}[ht]
  \centering
  \includegraphics[width=.43\textwidth]{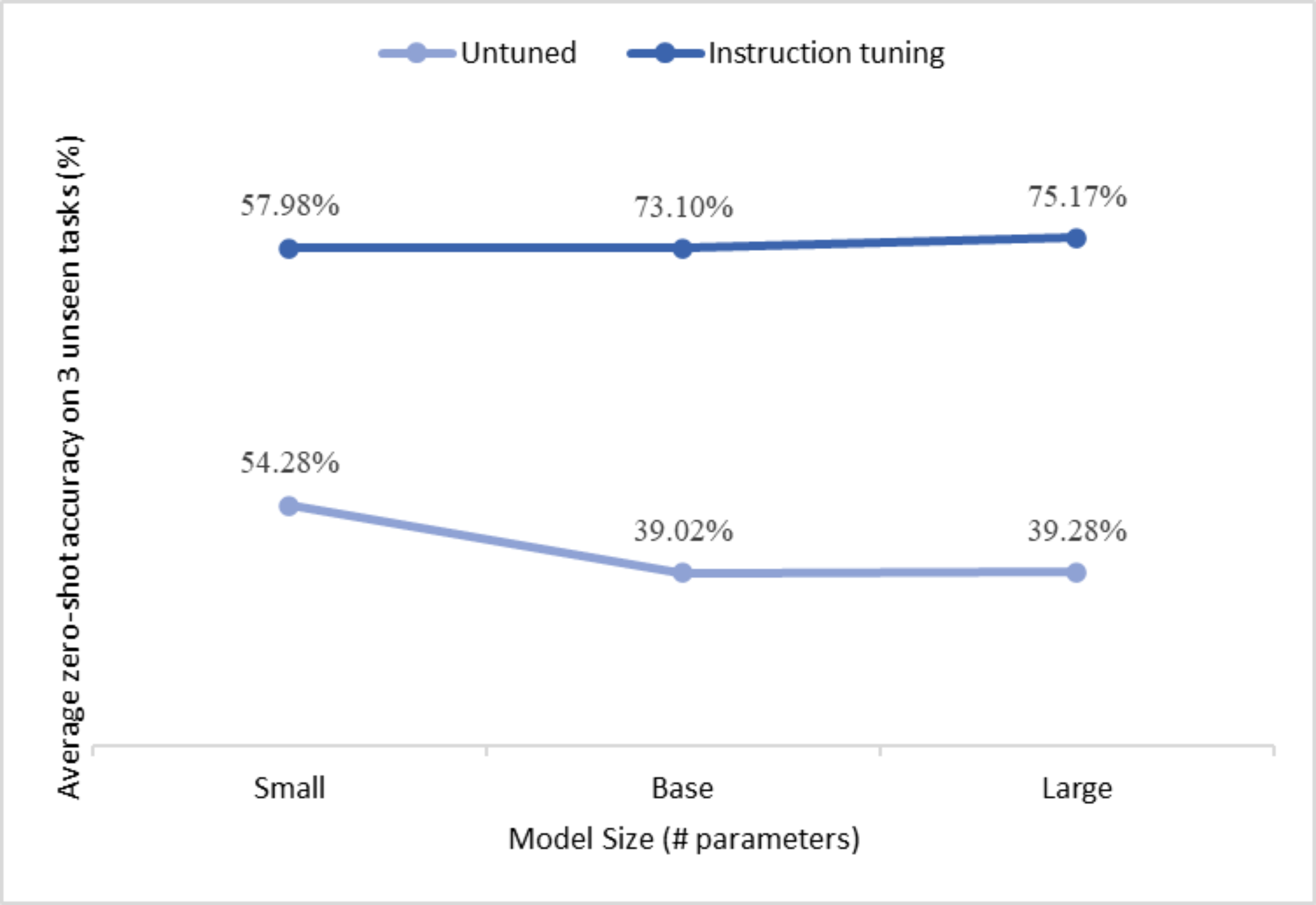}
  \caption{Evaluating instruction tuning efficacy for sentiment detection in FLAN-T5 models of varying sizes.}
  \label{fig: Performance analysis based on scaling laws}
\end{figure}

Table \ref{table_experiment for scaling law} shows the experimental results of classifying the crypto-related tweets for three models on three datasets. The models are further subdivided into two groups. One is the vanilla LM, and the other one is the instruction-based Fine-tuned model.
As depicted in Figure \ref{fig: Performance analysis based on scaling laws}, the results shed light on the effectiveness of instruction tuning with a larger model size enhancing the performance on unseen tasks. 
The untuned models achieved an average accuracy of 54.28\% for the small model, 39.02\% for the base model, and 39.28\% for the large model. 
\begin{table*}[!ht]
    \centering
    \caption{Architectural details of FLAN-T5 models.}
    \begin{adjustbox}{width=\textwidth}
    \begin{tabular}{cccccccc} 
        \hline
        Model & Architecture & \#Heads & \#Layers & Estimated model params size & \#Parameter & \#Trainable parameter & Model Checkpoint \\ 
        \hline
        FLAN-T5-SMALL & encoder-decoder & 6 & 8 & 308MB & 80M & 77M & google/flan-t5-small \\
        FLAN-T5-BASE & encoder-decoder & 12 & 12 & 990MB & 250M & 247M & google/flan-t5-base \\
        FLAN-T5-LARGE & encoder-decoder & 16 & 24 & 3133MB & 780MB & 783M & google/flan-t5-large \\
        \hline
    \end{tabular}
    \end{adjustbox}
\label{table_FLAN_architecture_detail}
\end{table*}
\begin{table*}[!ht]
    \centering
    \caption{Performance analysis of untuned and instruction-based Fine-tuned FLAN-T5 model across different sizes and datasets.}
    \begin{tabular}{cccc|ccc} 
        \hline
        \multirow{2}{*}{Model} & \multicolumn{3}{c}{Untuned} & \multicolumn{3}{c}{Instruction-based tuned} \\ 
        \cline{2-7}
         & Bitcoin sentiment & Reddit & Cryptocurrency sentiment & Bitcoin sentiment & Reddit & Cryptocurrency sentiment \\ 
        \hline
        FLAN-T5-SMALL & 59.98\% & 51.43\% & 51.43\% & 75.00\% & 49.46\% & 49.46\% \\
        FLAN-T5-BASE & 25.10\% & 46.07\% & 45.89\% & 75.00\% & 72.14\% & 72.14\% \\
        FLAN-T5-LARGE & 25.87\% & 46.07\% & 45.89\% & 76.94\% & 74.29\% & 74.29\% \\
        \hline
    \end{tabular}
\label{table_experiment for scaling law}
\end{table*}
The achieved accuracy result was inconsistent compared to the model's size. 
However, after applying instruction tuning, the accuracy improved to 57.98\% for the small model, 73.10\% for the base model, and 75.17\% for the large model.

This potential result could be explained based on the study conducted by Wei et al. [19] that instruction tuning serves two purposes for larger-scale models. 
Firstly, it occupies some of the model's capacity. 
Secondly, it instructs these models on following instructions, enabling them to apply this skill to new tasks using the remaining capacity, which helps large models generalize to new tasks.
But for small models, it actually hurts generalization to unseen tasks, potentially because all model capacity is used to learn the mixture of instruction tuning tasks.

\subsubsection{How does the instruction-based model respond over different instruction tuning setups?}
The conducted experimentation aimed to measure and compare the quality of models under different instruction tuning setups, focusing on the response of the instruction-based model. 
By introducing diversity in the styles and formats of tasks through instructions of varying lengths and complexities, the study provided insights into how the model handles diverse instructions and facilitates prompt tuning.
Generally, they can be formulated as the following equations:
\begin{equation}
    \text{{argmax}}_{M_{\text{{tuned}}}}\sum_{i\in I} Q(M_{\text{{tuned}}}, L_i, C_i)
\end{equation}

where $I = \{i_1, i_2, i_3, \ldots, i_n\}$.
 $Q(M_{\text{tuned}}, L_i, C_i)$ represents the quality of the instruction-tuned model.
$M_{\text{tuned}}$ be the instruction-tuned model.
$i_j$ represents the instruction.
$L_i$ denotes the length of instruction $i$.
$C_i$ denotes the complexity of instruction $i$.
Six instructions were created for this experiment, representing different lengths and complexities as shown in Table \ref{table_prompt_tuning}. 

\begin{figure}[ht]
  \centering
  \includegraphics[width=.43\textwidth]{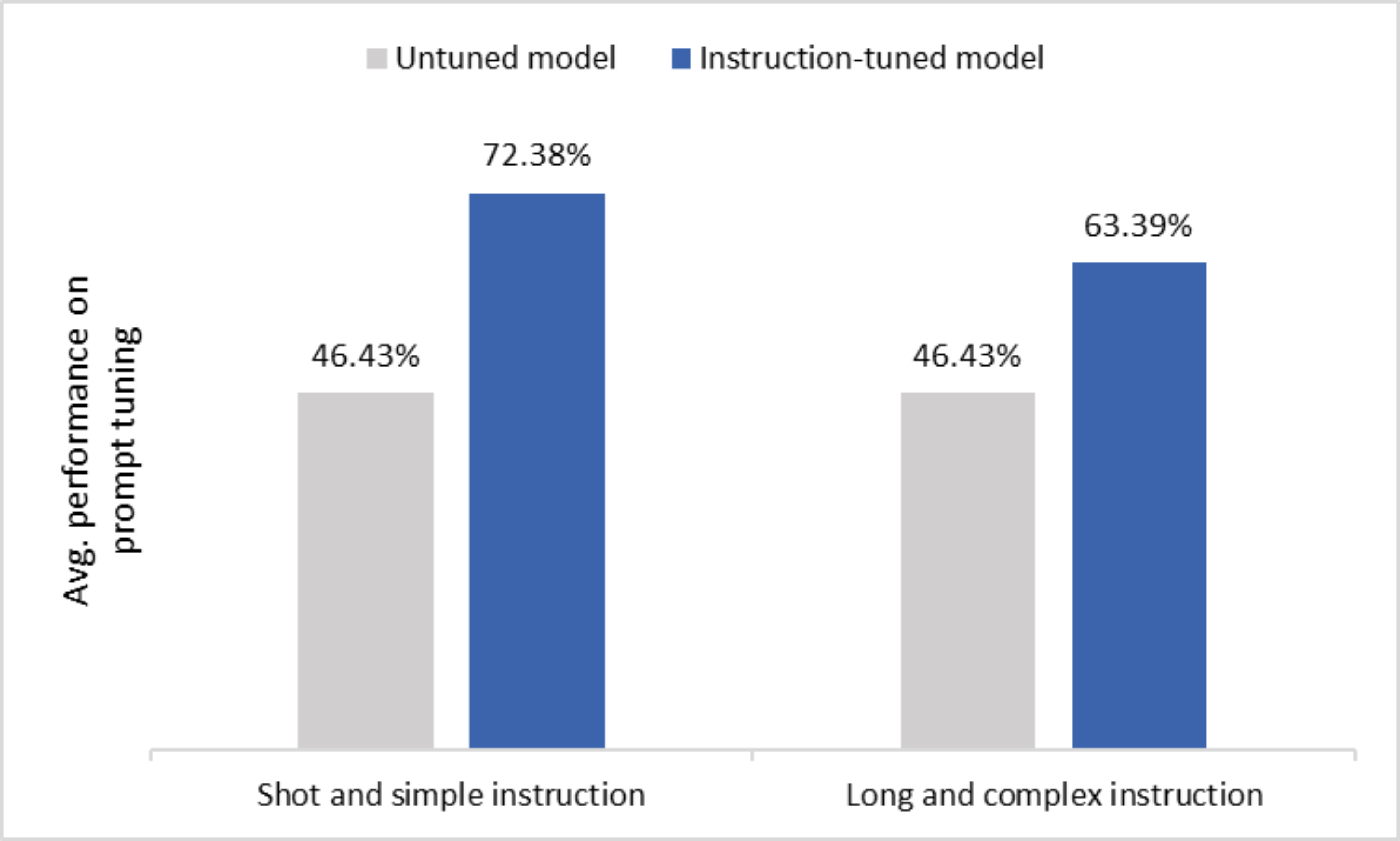}
  \caption{Performance analysis of untuned and instruction-based Fine-tuned FLAN-T5 model across various prompts.}
  \label{fig: Performance analysis based on prompt tuning}
\end{figure}

The results of prompt tuning experiments were analyzed for both short and simple instructions, as well as long and complex instructions, with the baseline being the FLAN-T5-BASE vanilla LM model as shown in Figure \ref{fig: Performance analysis based on prompt tuning}.
Preliminary experiments revealed that prompt tuning had no significant impact on the performance of the vanilla LM model, which achieved an average accuracy score of 46.43\% for both setups. 
However, prompt tuning demonstrated a significant improvement in the performance of the instruction-based model. Under short and simple instructions, the model achieved an average accuracy score of 72.38\%, showcasing its effectiveness in understanding and executing such instructions. 
On the other hand, the model exhibited slightly inferior performance for long and complex instructions, with an average accuracy score of 63.39\%.

The findings suggest that the instruction-based model excels in responding to and executing short and simple instructions, outperforming its performance under long and complex instructions by over 12\%. 

\begin{table}[!ht]
\caption{Prompt type variations for sentiment detection and instruction-based Fine-tuned model evaluation across diverse text types.}
\begin{tabular}{ll}
\hline
\multicolumn{1}{c}{Type}    & \multicolumn{1}{c}{Prompt}\\ \hline
\multirow{3}{*}{Shot and simple}     & \textit{Please detect the sentiment.} \\ \cline{2-2}
                            & \textit{Detect the sentiment of the text.}\\ \cline{2-2}
                            & \textit{Please detect the sentiment of the given text.}\\ \hline
\multirow{9}{*}{Long and complex} & \textit{\begin{tabular}[c]{@{}l@{}}Classify the sentiment of the provided cryptocurrency \\ related social media posts or messages.\end{tabular}}\\ \cline{2-2}
                            & \textit{\begin{tabular}[c]{@{}l@{}}Determine the emotional tone of the given text, which \\ primarily revolves around cryptocurrencies and their \\ associated concepts.\end{tabular}}\\ \cline{2-2}
                            & \textit{\begin{tabular}[c]{@{}l@{}}Categorize the sentiment expressed in the provided \\ dataset consisting of the text snippets related to \\ cryptocurrency and computer science, focusing on \\ capturing positive or negative sentiments.\end{tabular}} \\ \hline
\end{tabular}
\label{table_prompt_tuning}
\end{table}
\begin{table}[!ht]
    \centering
    \caption{Average zero-shot accuracy at different sample sizes on supervised Fine-tuned models.}
    \begin{tabular}{@{}cccccc@{}}
        \hline
        Sample size & DistilBert & MiniLM & FLAN-T5 & Average & Best Score \\ \hline
        2K & 54.58\% & 63.03\% & 58.51\% & 58.71\% & 63.03\% \\ 
        4K & 56.54\% & 69.08\% & 60.27\% & 61.93\% & 69.08\% \\ 
        \textbf{6K} & \textbf{66.64\%} & \textbf{69.67\%} & \textbf{61.16\%} & \textbf{65.82\%} & \textbf{69.67\%} \\ 
        8K & 57.83\% & 66.82\% & 60.89\% & 61.85\% & 66.82\% \\ 
        10K & 55.93\% & 54.58\% & 61.04\% & 57.19\% & 61.04\% \\ 
        12K & 58.26\% & 59.72\% & 60.93\% & 59.64\% & 60.93\% \\ 
        \hline
    \end{tabular}
\label{table_optimal_sample_size}
\end{table}
\subsubsection{How does the size of the fine-tuning dataset impact the performance of different language models, and what is the optimal sample size for achieving the highest performance of any model?}
The experimental evaluation aimed to explore the relationship between the size of the fine-tuning dataset and the SFT models' performance. 
Three models, namely DistilBert, MiniLM, and FLAN-T5-Base, were examined, and the sample size was varied from 2,000 to 12,000 data points to assess the influence of data availability. Table \ref{table_optimal_sample_size} presents the average zero-shot accuracy achieved by each model at different sample sizes on unseen tasks. 
The experimental findings provide valuable insights into the impact of fine-tuning dataset size on model performance, highlighting data efficiency and consistency considerations across the examined models. 
Among the three models, the best accuracy result was consistently achieved with a sample size of 6,000 data points. 
At this sample size, DistilBert reached an accuracy of 66.64\%, MiniLM achieved 69.67\%, and FLAN-T5-Base demonstrated an accuracy of 61.16\%. 
This indicates that 6,000 data points serve as an optimal balance for achieving the highest performance for each model. 
MiniLM exhibited significant data efficiency by achieving an average zero-shot accuracy of 63.03\% with a sample size of 2,000 data points. 
In comparison, FLAN-T5-Base and DistilBert required larger sample sizes, approximately double and triple the data, respectively, to attain comparable accuracy levels. 
This showcases MiniLM's ability to leverage limited data and deliver competitive performance effectively. 

Furthermore, FLAN-T5-Base displayed consistent performance across varying sample sizes, maintaining relatively stable average zero-shot accuracy values. 
This suggests the model's robustness and resilience to variations in the dataset size. 
Conversely, DistilBert exhibited performance fluctuations and a declining trend as the sample size increased beyond 6,000 data points. 
This implies that DistilBert may reach a saturation point where additional data does not contribute significantly to its performance improvement.

The findings have practical implications for model selection in the fine-tuning process. 
The data efficiency of MiniLM, combined with the consistent performance of FLAN-T5-Base, offers advantages in scenarios with limited data availability or where performance stability is crucial. 
Meanwhile, considering the appropriate sample size, such as 6,000 data points, is essential to optimize the performance of DistilBert.

\section{Conclusion}
In conclusion, the above research establishes the value of large language model fine-tuning strategies for sentiment analysis in the context of cryptocurrencies. Experimental results show a significant 40\% average gain in zero-shot performance after fine-tuning, highlighting the potential of this strategy in maximizing the effectiveness of pre-trained language models. Additionally, the result shows that instruction adjustment improves model performance, with larger models reaching a remarkable average accuracy score of 75.17\%. Notably, it has been found that 6,000 data points were the ideal corpus size, with MiniLM showing outstanding data efficiency and FLAN-T5 performing consistently across a range of corpus sizes. The research also shows that the model performs well when given short, clear instructions, surpassing its performance when given longer, more complicated instructions by almost 12\%. These observations help sentiment analysis in the cryptocurrency space improve while also offering helpful advice for large language model optimization. They should stimulate additional studies in supervised and instruction-based NLP, zero-shot learning, instruction tuning, and the use of labeled data to improve the effectiveness of big language models in cryptocurrency applications.

\bibliographystyle{unsrt}
\bibliography{library}

\EOD

\end{document}